%% file: main.tex
\documentclass[runningheads]{llncs}

\usepackage{eccv}

\input{preamble.tex}

\usepackage{hyperref}

\usepackage{orcidlink}

\begin{document}

\title{\ourmodel: Unified Dual-Modal Latent Diffusion for \\
Human-Centric Joint Video-Depth Generation} 

\titlerunning{IDOL for Human-Centric Joint Video-Depth Generation}

\author{Yuanhao Zhai\thanks{Work done during an internship at Microsoft.}\inst{1}\orcidlink{0000-0002-3277-3329} \and
Kevin Lin\inst{2}\orcidlink{0000-0001-8944-1336} \and
Linjie Li\inst{2}\orcidlink{0000-0003-0867-8863} \and
Chung-Ching Lin\inst{2}\orcidlink{0009-0003-6507-3657} \and
Jianfeng Wang\inst{2}\orcidlink{0000-0002-3156-4429} \and
Zhengyuan Yang\inst{2}\orcidlink{0000-0002-5808-0889} \and
David Doermann\inst{1}\orcidlink{0000-0003-1639-4561} \and
Junsong Yuan\inst{1}\orcidlink{0000-0002-7901-8793} \and
Zicheng Liu\inst{3}\orcidlink{0000-0001-5894-7828} \and
Lijuan Wang\inst{2}\orcidlink{0000-0002-5705-876X}}

\authorrunning{Y.~Zhai et al.}

\institute{State University of New York at Buffalo \and
Microsoft \and Advanced Micro Devices \\
\url{https://yhzhai.github.io/idol/}}

\maketitle

\input{sec/0_abstract}    
\input{sec/1_intro}
\input{sec/2_related_work}

\input{sec/3_method}
\input{sec/4_exp}
\input{sec/5_conclusion}
\input{sec/6_acknowledgement}

\bibliographystyle{splncs04}
\bibliography{main}

\appendix
\clearpage
\input{sec/1_exp}

\input{sec/2_details.tex}

\end{document}

%% file: preamble.tex
\usepackage{eccvabbrv}

\usepackage[accsupp]{axessibility}  %
\usepackage{graphicx}
\usepackage{amsmath}
\usepackage{amssymb}
\usepackage{microtype}
\usepackage{float}
\usepackage{subfloat}
\usepackage{bm}
\usepackage{multirow}
\usepackage{colortbl}
\usepackage{adjustbox}
\usepackage{enumitem}
\usepackage{rotating}
\usepackage{siunitx}
\usepackage{xspace}
\usepackage{booktabs}
\usepackage{wrapfig}

\newcommand{\ourmodel}{IDOL\xspace}
\newcommand{\fullmodelname}{unIfied Dual-mOdal Latent diffusion}
\definecolor{mygray}{gray}{0.9}  %
\usepackage[dvipsnames]{xcolor}

%% file: sec/0_abstract.tex
\begin{abstract}
Significant advances have been made in human-centric video generation, yet the
joint video-depth generation problem remains underexplored.
Most existing monocular depth estimation methods may not generalize well to
synthesized images or videos, and multi-view-based methods have difficulty
controlling the human appearance and motion.
In this work, we present \ourmodel (\fullmodelname) for high-quality
human-centric joint video-depth generation.
Our \ourmodel consists of two novel designs.
First, to enable dual-modal generation and maximize the information exchange
between video and depth generation, we propose a unified dual-modal U-Net, a
parameter-sharing framework for joint video and depth denoising, wherein a
modality label guides the denoising target, and cross-modal attention enables the
mutual information flow.
Second, to ensure a precise video-depth spatial alignment, we propose a motion
consistency loss that enforces consistency between the video and depth feature
motion fields, leading to harmonized outputs.
Additionally, a cross-attention map consistency loss is applied to align the
cross-attention map of the video denoising with that of the depth denoising,
further facilitating spatial alignment.
Extensive experiments on the TikTok and NTU120 datasets show our superior
performance, significantly surpassing existing methods in terms of video FVD and
depth accuracy.
\end{abstract}

%% file: sec/1_intro.tex
\section{Introduction}
\label{sec:intro}

The capacity to control and manipulate human-centric video content -- modulating
subjects' actions, altering appearance and background -- has attracted continual
study~\cite{siarohin2019animating,siarohin2021motion,ma2023follow,wang2023disco,ju2023humansd}.
With the rapid evolution of generative models, from generative adversarial
networks~\cite{brock2018large,goodfellow2020generative} to recent latent
diffusion models~\cite{rombach2022high}, the quality of the generated video has
improved significantly.
However, most of the existing
research~\cite{siarohin2019animating,siarohin2021motion,ma2023follow,wang2023disco,ju2023humansd}
has focused on the generation of 2D content.
This imposes a natural limit on applications that require depth perception, such
as virtual and augmented reality, as well as advanced video games.
In this paper, we explore joint video-depth generation for human actions (\eg{},
dancing and daily activities), where the video and the corresponding depth map are
simultaneously generated.
By learning a holistic representation of the human, it not only enhances the
visual fidelity of the synthesized content but also paves the way for
applications that demand a deeper spatial understanding.

\input{figures/fig-teaser.tex}

Existing methods confront several challenges in dealing with this task.
First, discriminative monocular depth estimation
methods~\cite{ranftl2020towards,jafarian2021learning}, which are typically trained on
natural images, have been empirically observed to exhibit degraded performance
when applied to generated
images~\cite{zheng2018t2net,atapour2018real,zhao2019geometry,bae2023deep}.
On the other hand, while multi-view-based
methods~\cite{watson2022novel,poole2022dreamfusion,gu2023nerfdiff,wynn2023diffusionerf,raj2023dreambooth3d}
can estimate depth, they primarily focus on the synthesis of individual
object/scene.
They often struggle with inference from single-view inputs or underperform in
manipulating the object's appearance and motion.
To address these problems, we propose to jointly generate the video and the corresponding depth.

However, joint video-depth generation presents nontrivial challenges for two
primary reasons.
First, video and depth are inherently two different modalities: the former is
represented as a 3-channel RGB frame sequence and the latter as a scalar depth
map sequence~\cite{ranftl2020towards,ranftl2021vision}.
Contrarily, prevailing diffusion models are pre-trained on the single-modal image
generation task~\cite{rombach2022high}.
Thus, designing a dual-modal diffusion model for joint video-depth generation is
challenging, not to mention harnessing the power of pre-trained latent diffusion
models.
Second, spatial layout control has been a long-standing problem in diffusion
models~\cite{hertz2022prompt,mokady2023null,tumanyan2023plug,geyer2023tokenflow}.
Even with human pose control, maintaining an accurate spatial alignment between
the generated video and depth remains a challenge.
This challenge can be more pronounced if the denoising process is conducted in
the latent space, given the intricate mapping from the latent space to the final
output.
In addressing these problems, we propose \ourmodel (\fullmodelname), a framework
that aims to generate a human-centric video and the corresponding depth jointly.

To tackle the problem of distinct video and depth representations, we propose to
render depth maps as RGB images by applying a color map to them.
This conversion reframes the depth generation task as a stylized video
generation problem.
Besides, existing methods suggest that incorporating depth as input enhances
structural understanding and boosts the generation
quality~\cite{esser2023structure}.
Drawing inspiration from this, we hypothesize that a richer interplay between
depth and video generation would be reciprocally beneficial.
To this aim, we design a unified dual-modal U-Net that shares parameters across
both video and depth denoising processes.
Our model leverages a modality label to specify the denoising target, \ie{},
video or depth.
In this way, it enables joint learning of depth and video for better generation
quality while remaining parameter-efficient.
Furthermore, a cross-modal attention layer is added during the joint denoising
process, enabling an explicit correlation between the joint video-depth
learning.

To ensure a precise video-depth alignment, we propose to synchronize the motion
pattern of the intermediate video and depth features.
Specifically, the intermediate video and depth features within the U-Net contain
semantic information, where regions with similar semantic meanings are
represented with similar
features~\cite{baranchuk2021label,tumanyan2023plug,geyer2023tokenflow}, as shown
in~\cref{fig:cost-volume}.
Thus, 
by using the proposed motion consistency loss,
we enforce a consistent
motion between video and depth features, thereby promoting a precise
video-depth spatial alignment.
Additionally, we propose a cross-attention map
consistency loss to further strengthen video-depth alignment. This is drawn inspiration from existing observations~\cite{hertz2022prompt,qi2023fatezero} on the influence of cross-attention maps on the spatial layout. Different from~\cite{hertz2022prompt,qi2023fatezero}, we consider spatial alignment across video and depth modalities.

In summary, our contributions are as follows.
\begin{itemize}
\item We propose \ourmodel, a pioneering framework for human-centric joint
video-depth generation.
Our \ourmodel features a parameter-sharing unified dual-modal U-Net for
video-depth denoising.
Besides, cross-modal attention modules are used to enable the mutual information
flow.
\item We propose a motion consistency loss and a cross-attention map consistency
loss to allow fine-grained video-depth spatial alignment.
\item We conduct extensive experiments on two distinct datasets: TikTok and
NTU120, whose depth maps were obtained using different
methods~\cite{ranftl2020towards,ranftl2021vision,jafarian2021learning}.
The results not only show a significant improvement in video quality and depth
accuracy over state-of-the-art methods, but also highlight the flexibility of
our method in adapting to different depth maps.
Furthermore, our experiments reveal that our \ourmodel can be easily adapted to
different diffusion models.
\end{itemize}

%% file: figures/fig-teaser.tex
\begin{figure*}[t!]
    \includegraphics[width=\textwidth]{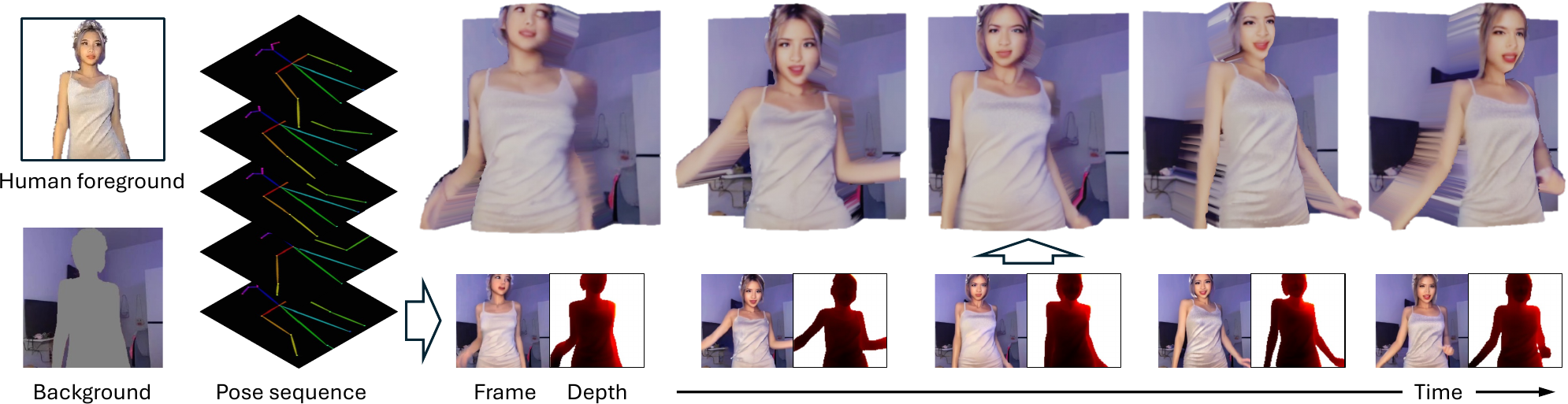}
    \caption{Given a human foreground image, an arbitrary background
    image, and a defined pose sequence, our \ourmodel generates high-fidelity
    video and the corresponding depth maps, which can be rendered as realistic 2.5D video.} 
    \label{fig:teaser}
\end{figure*}

%% file: sec/2_related_work.tex
\section{Related Work}
\label{sec:related-work}

\noindent\textbf{Controllable diffusion models.}
Evolving from early diffusion
models~\cite{ho2020denoising,dhariwal2021diffusion}, recent latent diffusion
models (LDM)~\cite{rombach2022high} achieves high-quality image generation by
conducting the denoising process in the latent space.
ControlNet~\cite{zhang2023adding}, GLIGEN~\cite{li2023gligen} and
T2I-Adapter~\cite{mou2023t2i} add trainable modules on pre-trained diffusion
models, and achieve fine-grained control given additional inputs, such as sketch
and depth.
HumanSD~\cite{ju2023humansd} directly concatenates the pose feature and the
noise in the input to avoid potential feature discrepancy problems.
In the video generation domain, controllable video diffusion
models~\cite{ma2023follow,chen2023control,hu2023videocontrolnet} leverage the
depth/flow/edge sequence to generate a temporally consistent video.
Our work builds upon these controllable models, utilizing plug-in modules to
facilitate human pose control in \ourmodel.
Our experiments demonstrate our adaptability across various frameworks (\eg{},
ControlNet~\cite{zhang2023adding} and T2I-Adapter~\cite{mou2023t2i}).

\noindent\textbf{Image animation.}
Image animation involves producing a video wherein an object from the source
image moves in congruence with the motion in a target video.
Traditional methods require specific knowledge about the target object, such
as facial landmarks~\cite{zakharov2019few,qian2019make,ha2020marionette},
gestures~\cite{tang2018gesturegan}, or semantic segmentation
maps~\cite{nirkin2019fsgan}.
Several methods learn a motion
field~\cite{wiles2018x2face,siarohin2019animating} from the driving videos.
First order motion model (FOMM)~\cite{siarohin2019animating} improves the
animation quality by learning a local affine transformation.
Motion representation for articulated animation (MRAA)~\cite{siarohin2021motion}
achieves better quality in representing articulated motions, such as the human
body.
Thin-plate spline (TPS)~\cite{zhao2022thin} estimates a more accurate motion,
and improves previous methods over large-scale motions.
Besides, there exists a line of  methods~\cite{reda2022film,mallya2022implicit}
exploiting end-to-end learning frameworks.
Based on diffusion models, DreamPose~\cite{karras2023dreampose} proposes an
adapter module, such as to control the appearance of the generated human.
DisCo~\cite{wang2023disco} disentangles the human attribute and pose condition
learning by employing a two-stage training scheme.
Several concurrent works improve DisCo~\cite{wang2023disco} using
more advanced appearance and motion
control~\cite{xu2023magicanimate,chang2023magicdance,hu2023animate}.
Furthermore, several
methods~\cite{ma2023follow,khachatryan2023text2video} inject motion/pose prior
to pre-trained text-to-image models.
Distinctively, our \ourmodel focuses on the joint video-depth generation, and utilizes depth generation to enhance the overall
quality of the animated video, instead
of video-only generation.

\noindent\textbf{Multi-modality synthesis.}
There exists a line of diffusion models for view synthesis, such as diffusion
models using multi-view
images~\cite{watson2022novel,gu2023nerfdiff,wynn2023diffusionerf}, point
cloud~\cite{nichol2022point}, and text-to-3D
models~\cite{poole2022dreamfusion,raj2023dreambooth3d}.
There are also methods developed for human body mesh
estimation~\cite{luan2021pc,luan2023high,luan2024spectrum}.
However, they may struggle with human-centric joint video-depth synthesis, as
they typically require multi-view input or lack the ability to precisely control
the appearance and motion of the target object/scene.
Except for the use of multi-view images, LDM3D~\cite{stan2023ldm3d} modifies the
autoencoder in LDM, such that the latent can be decoded into an RGB image and a
depth map.
MM-Diffusion~\cite{ruan2023mm} focuses on the audio-video generation, and
features a coupled U-Net structure to simultaneously denoise video and audio
latents.
Concurrently, HyperHuman~\cite{liu2023hyperhuman} proposes structural expert
branches to denoise latents of different modalities, such as RGB image, depth
and surface normal. 
To the best of our knowledge, there is no existing method directly working on
the joint video-depth generation task, and thus, we modify the backbones of
existing multi-modal generation methods for comparison.

\noindent\textbf{Diffusion models for dense prediction.}
Diffusion models have been used in various dense prediction tasks, such as
segmentation
tasks~\cite{baranchuk2021label,amit2021segdiff,wu2022medsegdiff,wolleb2022diffusion,chen2023generalist,ji2023ddp},
depth estimation~\cite{saxena2023monocular,ji2023ddp} and object
detection~\cite{chen2023diffusiondet}.
For depth estimation methods~\cite{saxena2023monocular,ji2023ddp}, they
typically necessitate modifying the output from 3-channel RGB images to scalar
depth maps.
Such a transformation impedes their ability to harness the power of large-scale
pre-trained diffusion models~\cite{rombach2022high}.
Concurrently, DepthAnything~\cite{depthanything} address this problem by
leveraging large-scale unlabeled data to improve the estimation quality.
Additionally, when applied to synthesized content, existing depth estimation
methods tend to
underperform~\cite{zheng2018t2net,atapour2018real,zhao2019geometry,bae2023deep},
failing to generate high-quality depth maps from outputs of image animation
methods.
To mitigate this problem, we propose \ourmodel to directly synthesize the video
and the corresponding depth, significantly improving the depth accuracy.

In contrast to existing methods, we reframe the depth synthesis task as a
stylized image synthesis task.
By rendering target depths as RGB images, we are positioned to directly use
pre-trained image generation models with minimal modifications.
Furthermore, we propose a unified dual-modal U-Net for improved joint
video-depth generation, and enhance video-depth spatial alignment via the
proposed consistency losses.

\input{figures/fig-structure.tex}

%% file: figures/fig-structure.tex
\begin{figure*}[t!]
    \includegraphics[width=\linewidth]{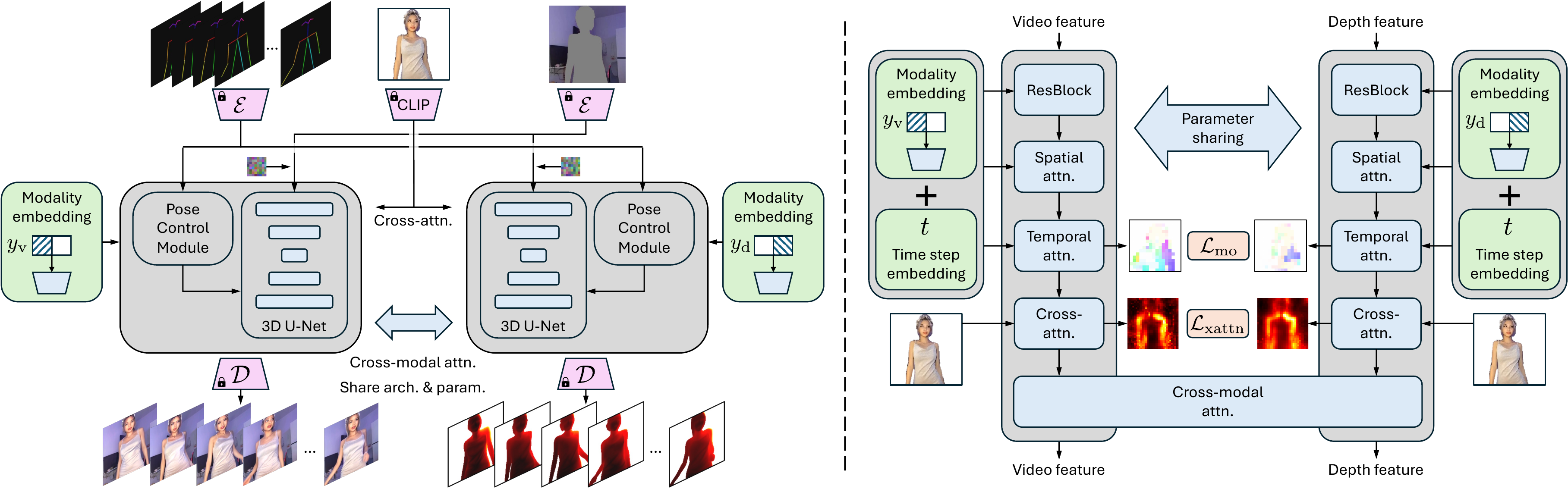}
    \caption{Left: \textbf{Overall model architecture.} Our \ourmodel features a unified
    dual-modal U-Net (gray boxes), a parameter-sharing design for joint video-depth
    denoising, wherein the denoising target is controlled by a one-hot modality label ($y_{\text{v}}$
    for video and $y_\text{d}$ for depth). Right: \textbf{Individual U-Net block structure.}
    Cross-modal attention is added to enable mutual information flow between video and
    depth features, with consistency loss terms $\mathcal{L}_{\text{mo}}$ and $\mathcal{L}_{\text{xattn}}$ ensuring the video-depth alignment. Skip
    connections are omitted for conciseness.}
    \label{fig:structure}
\end{figure*}

%% file: sec/3_method.tex
\section{Method}
\label{sec:method}

\noindent\textbf{Problem formulation.}
We begin by formally defining the task of human-centric joint video-depth generation.
Given a human foreground image ${f}$, a background image ${b}$, and a pose
sequence ${p} = \{{p}_1, {p}_2,..., {p}_L\}$ of length $L$, our objective is to
generate a video ${v} = \{ {v}_1, {v}_2, ..., {v}_L \}$ and the associated depth
map sequence ${d} = \{ {d}_1, {d}_2, ..., {d}_L \}$.
The video ${v}$ should faithfully animate the human foreground ${f}$ into the
target pose ${p}$ while integrating it with the background ${b}$.
The depth map sequence ${d}$ should correctly represent the depth within the
video ${v}$.
An illustration is shown in~\cref{fig:teaser}.

\noindent\textbf{Preliminaries.}
Latent diffusion models (LDM)~\cite{rombach2022high} recently show great
success in image generation.
It operates in the latent space of a pre-trained autoencoder
$\mathcal{D}(\mathcal{E}(\cdot))$, where a time-conditioned
U-Net~\cite{ronneberger2015u} $\epsilon_{\theta}(\cdot)$ with learnable
parameter $\theta$ is used for denoising the latent feature.
Within the U-Net, the conditional signal $c$ (\eg{}, textual prompt) is fed in
the cross-attention module after CLIP~\cite{radford2021learning} encoding.
During training, the objective is the mean square error (MSE) between the predicted and
ground truth noise:
\begin{equation}
\mathcal{L} = \mathbb{E}_{v, \epsilon, t, c} \left [ \| \epsilon -
\epsilon_{\theta}(z_t, t, c) \|_2^2 \right ],
\end{equation}
where $\epsilon \in \mathcal{N}(0, I)$ is the ground truth noise, and $z_t$ is
the noisy latent at the $t$-th reverse step, and can be obtained from
$\mathcal{E} (v)$ and $\epsilon$~\cite{ho2020denoising}.
To enable spatial layout control, ControlNet~\cite{zhang2023adding} adds a copy
of the U-Net down and middle blocks upon the LDM.
The output of ControlNet is added to the original U-Net via skip connections.
By manipulating the intermediate features, ControlNet trained with additional
conditional input (poses in our case) can control the output
image/video~\cite{wang2023disco}.

\subsection{Unified dual-modal U-Net}
\label{subsec:unified-unet}

Video and depth are different modalities and typically have distinct
representations.
Designing a model architecture that can jointly generate both video and depth,
and potentially leverage pre-trained diffusion models, presents a significant
challenge.
In addressing this problem, we propose to reformulate depth synthesis as a stylized
image generation task.
By rendering the depth map as a heatmap, we bridge the gap between the video and
depth modalities.
To further harness this approach, we introduce a unified dual-modal U-Net,
specially tailored for the joint video-depth denoising.

\noindent\textbf{Video LDM baseline.}
We build our model upon 3D U-Net~\cite{esser2023structure}, which modifies the
2D U-Net~\cite{rombach2022high} by adding a temporal convolutional layer in the 
ResBlock, and adding a temporal attention layer after the spatial attention
within each block.
For the human appearance and pose control, we follow DisCo~\cite{wang2023disco}
to feed the CLIP foreground latent $\text{CLIP}(f)$ via cross-attention, and
feed the pose latent $\mathcal{E}(p)$ via a ControlNet to the U-Net.
We show in~\cref{subsec:main-results} that our method can be implemented on
different pose control modules, such as T2I-Adapter~\cite{mou2023t2i}.
For the background control, different from DisCo that leverages a ControlNet, we
find that adding the background latent $\mathcal{E}(b)$ to the input noise of
the U-Net not only yields comparable results but also reduces the number of
parameters.
The proposed model architecture is illustrated in~\cref{fig:structure}.

\noindent\textbf{Sharing U-Net for joint video-depth denoising.}
Existing methods~\cite{esser2023structure} have shown that incorporating depth
map as conditional input enhances the model's video structure awareness,
improving the video generation quality.
Building on this insight, we hypothesize that maximizing the information
exchange between video and depth during generation can benefit the generation
quality for both modalities.
Thus, we propose a unified dual-modal U-Net for video and depth denoising, where
the architecture and parameters of the U-Net and ControlNet are shared between
the two modalities.
To indicate the denoising modality, we add a learnable modality embedding to
the time step embedding, which is fed to all blocks within the U-Net and the
ControlNet, as shown in \cref{fig:structure} right.
In this way, given a one-hot modality label, the corresponding modality
embedding will be selected to feed into the model and further control the output
modality.
Apart from being parameter-efficient, we find that this unified architecture
improves both video and depth generation quality (\cref{subsec:ablation}).

\noindent\textbf{Cross-modal attention.}
Though the unified dual-modal U-Net enables implicit structural information
learning, direct information communication during the video and depth denoising
process is needed for better alignment.
Thus, we add a multi-modality attention at the end of each block
(\cref{fig:structure} right).
During the joint video-depth denoising process, the video and depth features are
concatenated to conduct spatial self-attention.
Note that self-attention is only performed in the spatial dimension, as it aims
to promote spatial alignment between video and depth, and the preceding temporal
layers already ensure the temporal smoothness.

The joint video-depth denoising objective $\mathcal{L}_{\text{denoise}}$ is:
\begin{equation}
\begin{aligned}
\mathcal{L}_{\text{denoise}} = \mathbb{E}_{v, d, \epsilon_\text{v},
\epsilon_\text{d}, t, f, b, p} \big[ & \| \epsilon_\text{v} -
\epsilon_{\theta}(z_{\text{v},t}, t, f, b, p; y_{\text{v}}) \|_2^2 + \\
& \| \epsilon_\text{d} - \epsilon_{\theta}(z_{\text{d},t}, t, f, b, p;
y_{\text{d}}) \|_2^2 \big], \label{eq:overall-loss}
\end{aligned}
\end{equation}
where $y_\text{v}$ and $y_\text{d}$ are the modality labels for video and depth,
respectively, $\epsilon_{\text{v}}, \epsilon_{\text{d}} \in \mathcal{N}(0, I)$ are independently sampled Gaussian noises, and
$f, b, p$ denote the human foreground image, the background image, and the target pose sequence, respectively.
In~\cref{eq:overall-loss}, the first term is the video denoising loss and the
second term is the depth denoising loss.

\subsection{Learning video-depth consistency}
\label{subsec:learning-video-depth-consistency}

Though the pose control ensures the coarse correspondence between the video and
depth, we empirically find that only applying the joint denoising objective
$\mathcal{L}_{\text{denoise}}$ may still lead to misalignment between the
generated video and depth, especially when the training data is limited, as
shown in the first column of \cref{fig:cost-volume}.
To promote a precise video-depth alignment, we propose a motion consistency loss
$\mathcal{L}_{\text{mo}}$ and a cross-attention map consistency loss
$\mathcal{L}_{\text{xattn}}$.

\input{figures/comb-fig-cost-volume-pre-train.tex}

\noindent\textbf{Motion consistency loss.}
In image diffusion models, the intermediate self-attention features in the U-Net
are found to contain semantic
information~\cite{baranchuk2021label,tumanyan2023plug,geyer2023tokenflow}.
We found this finding also holds in our video-depth diffusion model.
However, though video and depth features share similar layouts, they may differ
in the temporal motion, as illustrated in ~\cref{fig:cost-volume}.
Thus, we attribute the video-depth misalignment to the unsynchronized motion
between video and depth features.
To address this problem, we propose to enforce a synchronized motion between the
video and depth features via a motion consistency loss
$\mathcal{L}_{\text{mo}}$.

Specifically, given intermediate video self-attention feature maps at frame $l$ and
$l+1$: $F_{\text{v},l}, F_{\text{v},l+1} \in \mathbb{R}^{H\times W\times D}$,
their cost volume $C_{\text{v},l}\in \mathbb{R}^{H\times W\times H\times W}$ can
be constructed by conducting cosine similarity between all point pairs on the
two frames:
\begin{equation}
c_{\text{v},l,i,j,h,k} = f_{\text{v},l,i,j} \cdot f_{\text{v},l+1,h,k},
\end{equation}
where $f_{\text{v},l,i,j}$ is the normalized feature vector on $F_{\text{v},l}$
at spatial location $i,j$.
The cost volume $C_{\text{d},r}$ for the depth feature maps can be computed
in a similar way.
As the video and depth features may distributed differently, we further
normalize the cost volume via softmax, resulting in a motion field
$U_{\text{v},l}$:
\begin{equation}
u_{\text{v},l,i,j,h,k} = \frac{\exp{(c_{\text{v},l,i,j,h,k} / \tau)}}{\sum_{h'}
\sum_{k'} \exp{(c_{\text{v},l,i,j,h',k'} / \tau)}},
\end{equation}
where $\tau$ is a temperature hyper-parameter controlling the concentration of
the distribution.
A high value on the motion field indicates a movement between two points.
The motion consistency is achieved by minimizing the MSE loss between
the video and depth motion fields:
\begin{equation}
\mathcal{L}_{\text{mo}} = \frac{1}{LHWHW} \sum_{l,i,j,h,k} \|
u_{\text{v},l,i,j,h,k} - u_{\text{d},l,i,j,h,k} \|_2^2.
\end{equation}
By enforcing similar motion of video and depth features, it promotes a
consistent video and depth output.

\noindent\textbf{Cross-attention map consistency loss.}
Except for leveraging the self-attention feature, the cross-attention maps
between the foreground image and the input noise are also found to be critical
in the image layout control~\cite{hertz2022prompt,qi2023fatezero}.
Thus, we additionally enforce the cross-attention maps from the video stream and
the depth stream to be similar via an MSE loss
$\mathcal{L}_{\text{xattn}}$:
\begin{equation}
\mathcal{L}_{\text{xattn}} = \| M_{\text{v}} - M_{\text{d}} \|_2^2,
\end{equation}
where $M_\text{v}$ and $M_\text{d}$ represent the video and depth cross-attention map,
respectively.

As the ControlNet fuses features in the U-Net up blocks, we apply the
consistency loss terms only to the U-Net up blocks to efficiently learn the
U-Net and the ControlNet.
The overall training objective $\mathcal{L}$ is defined as: \begin{equation}
\mathcal{L} = \mathcal{L}_{\text{denoise}} + \sum_{n=1}^{N} \left(w_{\text{mo}}
\mathcal{L}_{\text{mo},n} + w_{\text{xattn}} \mathcal{L}_{\text{xattn},n}
\right), \end{equation}
where $w_{\text{mo}}$ and $w_{\text{xattn}}$ are weighting hyper-parameters, $N$
is the number of up blocks, and $\mathcal{L}_{\text{mo},n}$ and
$\mathcal{L}_{\text{xattn},n}$ are the motion consistency loss and the
cross-attention map consistency loss on the $n$-th up block, respectively.

\subsection{Human attribute outpainting pre-training}
\label{subsec:pretrain}

To disentangle the pose and background control, DisCo~\cite{wang2023disco}
proposes a human attribute pre-training (HAP) task to learn human appearance.
Specifically, HAP learns to reconstruct the whole image given human foreground
and background images, with the pose ControlNet being removed, as shown
in~\cref{fig:pre-train} top.
However, we empirically find that such pre-training may lead to an apparent
background mask when the target pose deviates from the original position.

To address this problem, we propose human attribute outpainting pre-training
(HAOP).
Our HAOP differs from HAP in two ways.
First, HAOP extends the background mask through dilation, forcing the model to
fill in the background around the foreground.
Second, we apply a random crop and resize on the foreground image, prompting the
model to extrapolate the partial human attributes.
This process not only mitigates the background masking problem but also improves
image synthesis, as demonstrated in \cref{fig:pre-train} and validated in our
ablation studies (\cref{subsec:ablation}).
Formally, the pre-training learning objective is as follows:
\begin{equation}
\mathcal{L}_{\text{HAOP}} = \mathbb{E}_{v, \epsilon , t, f_{\text{aug}},
b_{\text{aug}}} \big[ \| \epsilon - \epsilon_{\theta}(z_t, t, f_{\text{aug}},
b_{\text{aug}}) \|_2^2 \big],
\end{equation}
where $f_{\text{aug}}$ and $b_{\text{aug}}$ represent augmented foreground and
background, respectively.
Note that the pre-training does not require depth denoising, which significantly
lowers the data demand.

%% file: figures/comb-fig-cost-volume-pre-train.tex
\begin{figure}[t]
    \centering
    \begin{minipage}{.5\textwidth}
        \includegraphics[width=\linewidth]{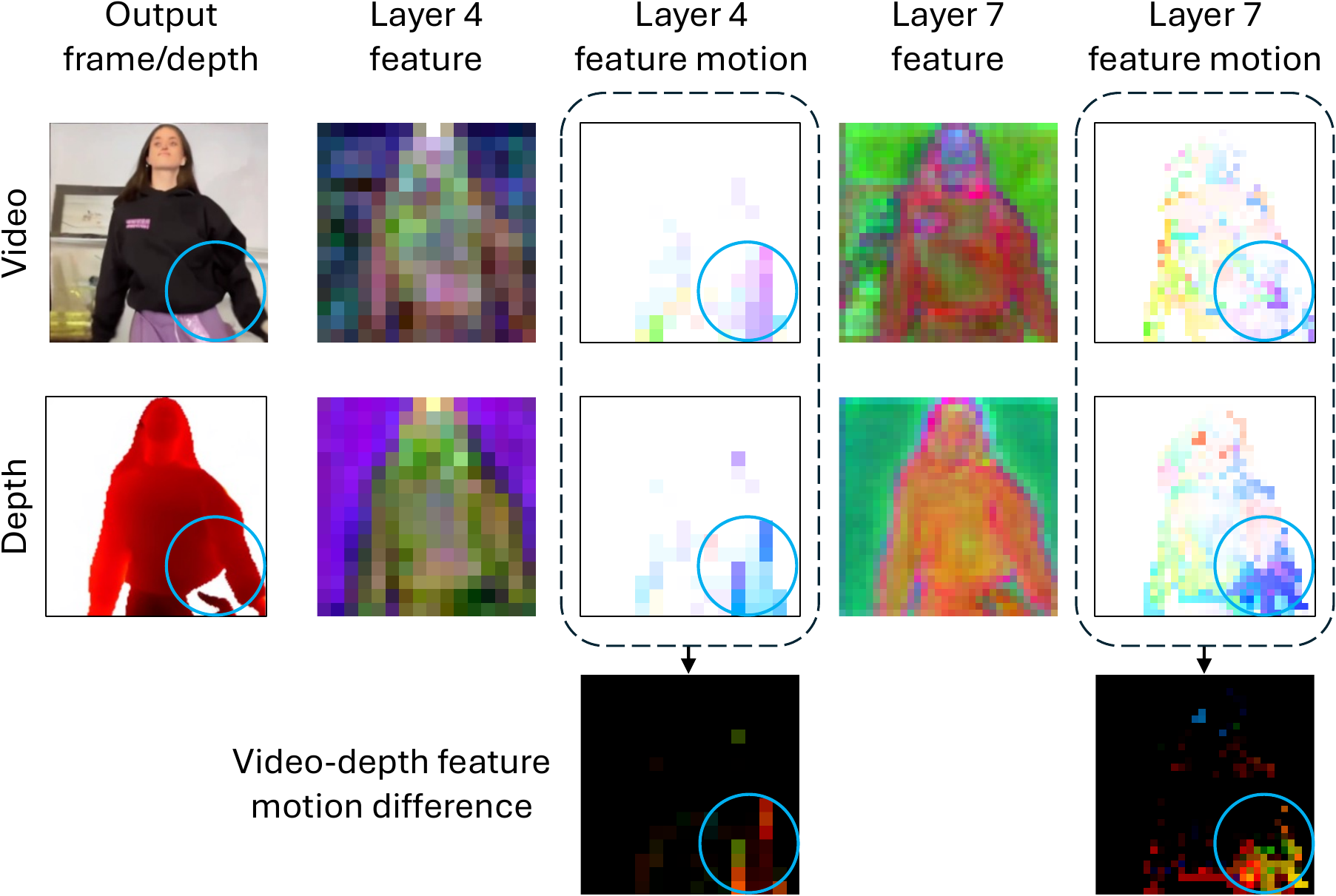}
        \caption{%
        Visualization of the video and depth feature maps and their motion fields without consistency losses.
        We attribute the inconsistent video-depth output (blue circle) to the
        inconsistent video-depth feature motions (the last row).
        This problem exists in multiples layers within the U-Net, and we
        randomly select two layers for visualization.
        We follow~\cite{tumanyan2023plug} to visualize the feature maps, and
        different color in the motion filed indicates different moving
        direction.
        }
        \label{fig:cost-volume}
    \end{minipage}%
    \hfill
    \begin{minipage}{.48\textwidth}
        \includegraphics[width=\linewidth]{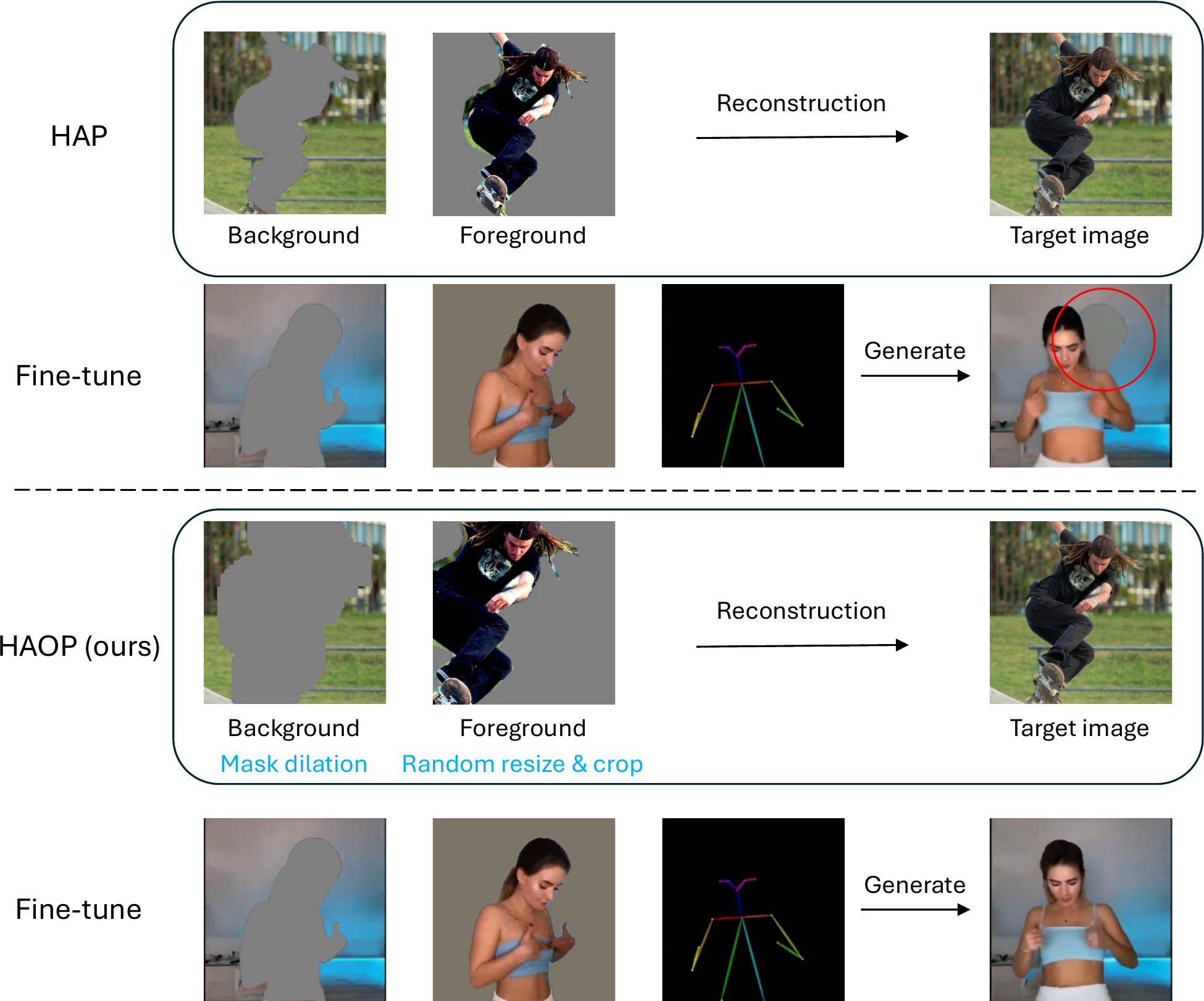}
        \caption{Comparison between human attribute pre-training
        (HAP)~\cite{wang2023disco} and our human attribute outpainting pre-training
        (HAOP). HAP may result in an apparent background mask when the target pose
        deviates from the original position, while our HAOP mitigates this problem.}
        \label{fig:pre-train}
    \end{minipage}
\end{figure}

%% file: sec/4_exp.tex
\section{Experiments}
\label{sec:exp}

\noindent\textbf{Datasets.}
We train and evaluate on two public datasets: TikTok~\cite{jafarian2021learning}
and NTU120~\cite{shahroudy2016ntu,liu2019ntu}.
The TikTok dataset consists of $\sim$350 human dancing videos.
We follow DisCo~\cite{wang2023disco} to use 335 videos for training, and use 10
videos for evaluation.
The NTU120 dataset consists of daily activities videos, we select 588 videos for
training and 72 videos for evaluation.
The videos are cropped to center to the subject.
We specifically ensure distinct subjects and backgrounds between training and
evaluation, and select only certain subjects to better evaluate the
generalization ability.
Besides, we use Grounded-SAM~\cite{kirillov2023segment,liu2023grounding} for the
human foreground mask estimation, and follow ControlNet~\cite{zhang2023adding}
to use OpenPose~\cite{cao2017realtime} for human pose estimation.
For the depth estimation, we leverage two different methods:
HDNet~\cite{jafarian2021learning} for high-fidelity depth estimation on the
human foreground, where the background depth is set to a constant;
MiDaS~\cite{ranftl2020towards,ranftl2021vision} for the whole-frame depth
estimation, which lacks details on the human foreground.
The HDNet depth map is rendered by applying the ``hot'' colormap; the MiDaS
depth map is rendered as grayscale images.
For pre-training, we follow DisCo~\cite{wang2023disco} to use $\sim$700k images
from a combined dataset of TikTok~\cite{jafarian2021learning},
COCO~\cite{lin2014microsoft}, SHHQ~\cite{fu2022stylegan},
DeepFashion2~\cite{ge2019deepfashion2}, and LAION~\cite{schuhmann2021laion}.

\noindent\textbf{Evaluation metrics.}
We separately evaluate the quality of the generated video and depth.
For video quality evaluation, we follow DisCo~\cite{wang2023disco} to use
FID-FVD~\cite{balaji2019conditional} and FVD~\cite{unterthiner2018towards} over
every 8-frame snippet.
For the evaluation of depth synthesis, for simplicity, we first scale the depth in the
range $[0, 1]$, and compute L2 distance between the ground truth
depth\footnote{The depth evaluation is conducted on the synthesized depth and
the depth estimated from the ground truth image instead of that estimated from
the generated images. This is because we empirically find that the depth estimated from the
generated image tends to be noisy (see \cref{fig:qa-comparison-hdnet} and
\cref{fig:qa-comparison-midas}).} and the generated depth.
In addition to evaluating the generated video and depth, we further use 
FID~\cite{heusel2017gans} on the frames to measure the image quality.

\noindent\textbf{Comparison methods.}
We adopt state-of-the-art image animation methods
FOMM~\cite{siarohin2019animating}, MRAA~\cite{siarohin2021motion},
TPS~\cite{zhao2022thin}, DreamPose~\cite{karras2023dreampose}, and
DisCo~\cite{wang2023disco} for the video and image quality comparison.
As a pioneering method in human-centric joint video-depth generation, to our
knowledge, there is no direct prior work for the joint video-depth generation
quality comparison.
Thus, we adapt the most relevant multi-modal generation methods
(LDM3D~\cite{stan2023ldm3d} for text to image-depth and
MM-Diffusion~\cite{ruan2023mm} for text to video-audio), modify their backbones
to the same VideoLDM baseline as ours (\cref{subsec:unified-unet}) to enable
appearance and pose control, and compare the joint video-depth generation results with
them.

\input{tables/tab-comp-with-sota-tiktok.tex}

\subsection{Main results}
\label{subsec:main-results}

In our experiments, we use the HDNet depth~\cite{jafarian2021learning} by
default for its high-fidelity human foreground depth estimation.
We also train and evaluate on the MiDaS
depth~\cite{ranftl2020towards,ranftl2021vision} to demonstrate the generalization ability of our
\ourmodel.

\input{figures/fig-qualitative-hdnet.tex}

\input{figures/comb-fig-comp-on-midas-qualitative-midas.tex}

\noindent\textbf{Comparison with the state-of-the-art.}
We compare \ourmodel with state-of-the-art methods on the TikTok and NTU120
datasets in~\cref{tab:comp-with-sota-tiktok-ntu-hdnetdepth}, with the generated
frame and depth visualized in~\cref{fig:qa-comparison-hdnet}.
We make the following observations.
(1)~We outperform all competing motion transfer and video generation
methods~\cite{siarohin2019animating,siarohin2021motion,zhao2022thin,karras2023dreampose,wang2023disco}
in terms of both video and image quality.
(2)~When adapting multi-modal generation
methods~\cite{stan2023ldm3d,ruan2023mm} for human-centric joint video-depth
generation, with the integration of the video LDM baseline
(\cref{subsec:unified-unet}), \ourmodel consistently outperforms them across
video, image, and depth metrics on both datasets, demonstrating the effectiveness of our method.
(3) Applying HDNet depth estimation to the previous state-of-the-art DisCo
generated frames leads to suboptimal results, with incomplete, fragmentary or
overly simplistic estimations, as shown in~\cref{fig:qa-comparison-hdnet}.
As the estimation result deviates from the ground truth, the HDNet results lead
to a high depth L2, as listed in the DisCo row
of~\cref{tab:comp-with-sota-tiktok-ntu-hdnetdepth}.
This finding indicates the limited generalization ability of monocular depth
methods, and highlights the need for multi-modal generation approaches.

\noindent\textbf{Comparison on synthesizing different types of depth map.}
\cref{tab:comp-on-midasdepth} compares the results on using the MiDaS grayscale
depth map~\cite{ranftl2020towards,ranftl2021vision} for the joint video-depth
generation.
The results consistently show that our \ourmodel outperforms existing multi-modal
generation methods~\cite{stan2023ldm3d,ruan2023mm} in both video and depth
quality.
This further highlights \ourmodel's effectiveness and its adaptability to
different types of depth maps.
Additionally, as shown in the DisCo row of~\cref{fig:qa-comparison-midas}, applying MiDaS on the DisCo
generated frames also yields suboptimal results, blurring the areas surrounding the
human subject, and leads to poorer depth accuracy (measured by L2 norm) than that
achieved by generative approaches.

\input{tables/tab-adaptation.tex}

\noindent\textbf{Generalization ability.}
Our \ourmodel follows DisCo~\cite{wang2023disco} for setting the
default motion representations (OpenPose~\cite{cao2017realtime}) and backbone
architectures (ControlNet~\cite{zhang2023adding}).
In~\cref{tab:adaptation} and \cref{fig:adaptation}, we further show that our
\ourmodel can be easily adapted to other motion representations, temporal
modeling modules, and pose control modules.
First, \ourmodel can benefit from the motion prior from pre-trained
AnimateDiff~\cite{guo2023animatediff} to improve the video quality.
Besides, our \ourmodel can be conditioned to more fine-grained
DWPose~\cite{yang2023effective}, which provides additional facial and hand
keypoints, to improve the video and depth quality.
Impressively, even when using OpenPose, which lacks hand keypoint as condition,
\ourmodel effectively generates plausible human hands,
demonstrating its robust human attribute learning capabilities.
Furthermore, our \ourmodel can be implemented with different pose control
modules, such as ControlNet~\cite{zhang2023adding} and
T2I-Adapter~\cite{mou2023t2i}, achieving similar quantitative and qualitative results.
Such results demonstrate that our \ourmodel can be generalized to different design
options.

\input{figures/comb-fig-qa-ablation-tab-ablation-on-pre-train.tex}

\input{tables/comb-tab-ablation-on-architecture-ablation-on-loss.tex}

\subsection{Ablation studies}
\label{subsec:ablation}

We conduct a set of ablation studies on the TikTok
dataset~\cite{jafarian2021learning} with HDNet depth~\cite{jafarian2021learning}
to demonstrate the effectiveness of our proposed method.
These studies are conducted accumulatively, layering each component to assess
its incremental impact on the overall performance.

\noindent\textbf{Unified dual-modal U-Net for joint video-depth denoising.}
We analyze whether joint video-depth learning improves the performance, and the
effectiveness of the designs proposed in the unified dual-modal U-Net
in~\cref{tab:ablation-on-architecture}.
The key observations are as follows.
(1)~Jointly video-depth learning with a shared U-Net is beneficial for both
video and depth generation (the second row), while using only half the
parameters compared to the separate counterpart.
Such a result underscores the significance of our structurally-aware shared
U-Net.
(2)~The explicit cross-modal information exchange between the video and depth
denoising (the third row), \ie, through cross-modal attention, further improves
both the video and depth generation quality.
Moreover, as shown in \cref{fig:qa-ablation}, our unified dual-modal U-Net
markedly improves frame and depth quality compared to the separate U-Net baseline.
Such results confirm the effectiveness of our design.

\noindent\textbf{Learning video-depth consistency.}
In our \ourmodel, we introduce a motion consistency loss
$\mathcal{L}_{\text{mo}}$ and a cross-attention map consistency loss
$\mathcal{L}_{\text{xattn}}$ to enhance video-depth alignment.
We analyze their impact in \cref{tab:ablation-on-losses}, where the results
demonstrate that both of them improve the depth generation quality, and the motion
consistency loss improves all video, depth, and image generation results.
The final column of \cref{fig:qa-ablation} further illustrates how these
consistency loss terms effectively refine video-depth alignment in our model.

\input{tables/comb-tab-ablation-on-pretrain-complexity.tex}

\noindent\textbf{Human attribute outpainting pre-training.}
We analyze the effectiveness of our proposed human attribute outpainting
pre-training (HAOP) in~\cref{tab:comp-on-pre-train}.
The results reveal that HAOP surpasses human attribute pre-training
(HAP)~\cite{wang2023disco} across all metrics, confirming its efficacy.
Notably, even without pre-training, our \ourmodel already outperforms other
competing multi-modal generation methods~\cite{ruan2023mm,stan2023ldm3d}.

\noindent\textbf{Complexity and computational requirements.}
We analyzed the complexity of generating an 8-frame video-depth sequence on a
single V100 GPU, comparing our method with other multi-modal generation methods in~\cref{tab:complexity}.
Thanks to our unified U-Net design, \ourmodel requires the fewest FLOPs, has the
lowest number of trainable parameters, and the shortest inference time,
demonstrating our efficiency.

%% file: tables/tab-comp-with-sota-tiktok.tex
\begin{table*}[t!]
\centering
\resizebox{\linewidth}{!}{%
\begin{tabular}{c|c|cccc|cccc}
\hline 
\multirow{3}{*}{Method} & \multirow{3}{*}{Motion control} & \multicolumn{4}{c|}{{TikTok}} & \multicolumn{4}{c}{NTU120}\tabularnewline
 &  & \multicolumn{2}{c}{{Video}} & \multicolumn{1}{c}{{Depth}} & {Image} & \multicolumn{2}{c}{Video} & \multicolumn{1}{c}{Depth} & Image\tabularnewline
 &  & {FID-FVD$\downarrow$} & {FVD$\downarrow$} & L2$\downarrow$ & {FID$\downarrow$} & {FID-FVD$\downarrow$} & {FVD$\downarrow$} & L2$\downarrow$ & {FID$\downarrow$}\tabularnewline
\hline 
FOMM~\cite{siarohin2019animating} & \multirow{3}{*}{Target video} & 38.36 & 404.31 & {-} & 85.03 & 40.34 & 1439.50 & {-} & 80.29\tabularnewline
MRAA~\cite{siarohin2021motion} &  & 24.11 & 306.49 & {-} & 54.47 & 58.19 & 1441.79 & {-} & 97.07\tabularnewline
TPS~\cite{zhao2022thin} &  & 29.20 & 337.79 & {-} & 53.78 & 37.42 & 1339.86 & {-} & \underline{61.75}\tabularnewline
\cline{2-2} 
DreamPose~\cite{karras2023dreampose} & DensePose~\cite{guler2018densepose} & 52.62 & 614.07 & {-} & 75.08 & 80.11 & 791.25 & {-} & 116.23\tabularnewline
\cline{2-2} 
DisCo~\cite{wang2023disco} & \multirow{3}{*}{OpenPose~\cite{cao2017realtime}} & \underline{20.75} & \underline{257.90} & 0.0975$^\dagger$ & \underline{39.02} & \underline{26.21} & \underline{458.92} & \underline{0.0371}$^\dagger$ & 68.53\tabularnewline
LDM3D~\cite{stan2023ldm3d} &  & 45.30 & 553.03 & 0.0637 & 69.36 & 71.11 & 587.84 & 0.0650 & 120.74\tabularnewline
MM-Diffusion~\cite{ruan2023mm} &  & 48.92 & 771.32 & \underline{0.0367} & 68.47 & 58.44 & 504.05 & 0.0404 & 102.77\tabularnewline
\hline 
\ourmodel & OpenPose~\cite{cao2017realtime} & \textbf{17.86} & \textbf{223.69} & \textbf{0.0336} & \textbf{36.04} & \textbf{20.23} & \textbf{314.82} & \textbf{0.0317} & \textbf{50.70}\tabularnewline
\hline 
\end{tabular}%
}
\caption{Quantitative results comparison between our \ourmodel and existing methods on
the TikTok and NTU120 datasets. The ground truth depth is estimated via
HDNet~\cite{jafarian2021learning}. ``-'' indicates incapable of generating depth
map, ``$\dagger$'' indicates the HDNet-inferred depth from the synthesized image. The best and the second best results are denoted in \textbf{bold} and \underline{underscored}, respectively.}
\label{tab:comp-with-sota-tiktok-ntu-hdnetdepth}
\end{table*}

%% file: figures/fig-qualitative-hdnet.tex
\begin{figure*}[t!]
    \includegraphics[width=\linewidth]{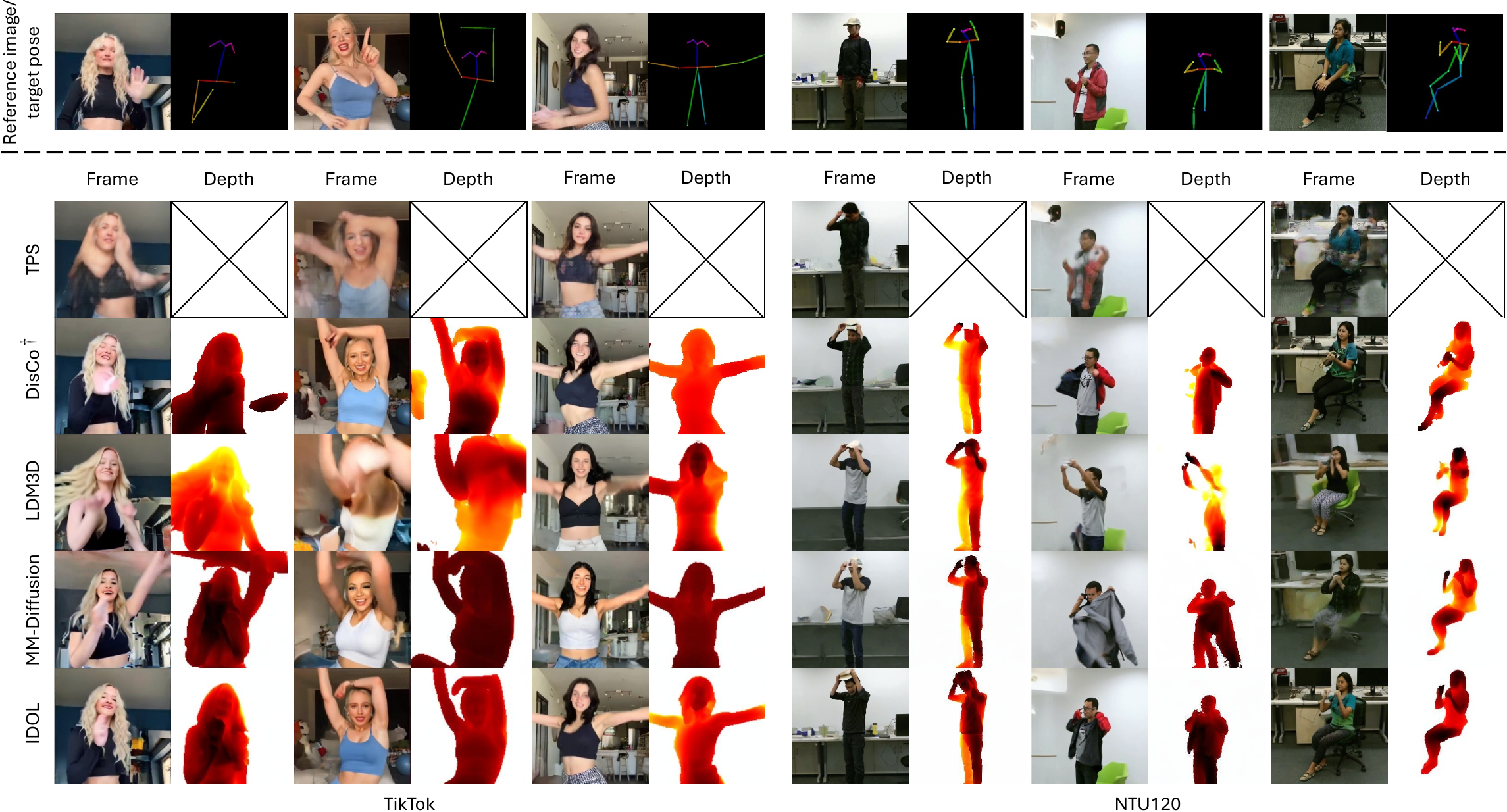}
    \caption{Qualitative results comparison between TPS~\cite{zhao2022thin},
    DisCo~\cite{wang2023disco}, LDM3D~\cite{stan2023ldm3d}, 
    MM-Diffusion~\cite{ruan2023mm} and our \ourmodel on the TikTok and NTU120 datasets with HDNet
    estimated depth~\cite{jafarian2021learning}. Note that
    TPS~\cite{zhao2022thin} is unable to generate depth, and the depth map for
    DisCo is estimated via HDNet on its generated frames. Please find video
    comparison in the supplementary material.}
    \label{fig:qa-comparison-hdnet}
\end{figure*}

%% file: figures/comb-fig-comp-on-midas-qualitative-midas.tex
\begin{figure}[t]
\begin{minipage}[h]{0.48\linewidth}
    \centering
    \resizebox{\linewidth}{!}{%
    \begin{tabular}{cccccc}
    \hline 
    \multirow{2}{*}{} & \multirow{2}{*}{Method} & \multicolumn{2}{c}{{Video}} & \multicolumn{1}{c}{{Depth}} & \multicolumn{1}{c}{{Image}}\tabularnewline
     &  & {FID-FVD$\downarrow$} & {FVD$\downarrow$} & {L2$\downarrow$} & {FID$\downarrow$}\tabularnewline
    \hline 
    \multirow{4}{*}{\begin{turn}{90}
    TikTok
    \end{turn}} & DisCo~\cite{wang2023disco} & \underline{20.75} & \underline{257.90} & 0.1758$^\ddagger$ & \textbf{39.02}\tabularnewline
     & LDM3D~\cite{stan2023ldm3d} & 42.09 & 529.43 & \underline{0.0646} & 72.30\tabularnewline
     & MM-Diffusion~\cite{ruan2023mm} & 52.37 & 715.28 & 0.1040 & 70.09\tabularnewline
    \cline{2-6} \cline{3-6} \cline{4-6} \cline{5-6} \cline{6-6} 
     & \ourmodel & \textbf{19.01} & \textbf{216.96} & \textbf{0.0271} & \underline{39.76}\tabularnewline
    \hline 
    \multirow{4}{*}{\begin{turn}{90}
    NTU120
    \end{turn}} & DisCo~\cite{wang2023disco} & \underline{26.21} & \underline{458.92} & 0.0695$^\ddagger$ & \underline{68.53}\tabularnewline
     & LDM3D~\cite{stan2023ldm3d} & 77.04 & 591.03 & \underline{0.0244} & 115.16\tabularnewline
     & MM-Diffusion~\cite{ruan2023mm} & 73.47 & 503.53 & 0.0260 & 117.22\tabularnewline
    \cline{2-6} \cline{3-6} \cline{4-6} \cline{5-6} \cline{6-6} 
     & \ourmodel & \textbf{20.56} & \textbf{439.63} & \textbf{0.0210} & \textbf{55.70}\tabularnewline
    \hline 
    \end{tabular}%
    }
    \captionof{table}{Quantitative comparison between our method and existing methods on the
    TikTok and NTU120 datasets with MiDaS estimated
    depth~\cite{ranftl2020towards,ranftl2021vision}. ``$\ddagger$'' indicates the
    MiDaS estimated depth from the synthesized image.}
    \label{tab:comp-on-midasdepth}
\end{minipage}
\hfill
\begin{minipage}[h]{0.5\linewidth}
    \includegraphics[width=\linewidth]{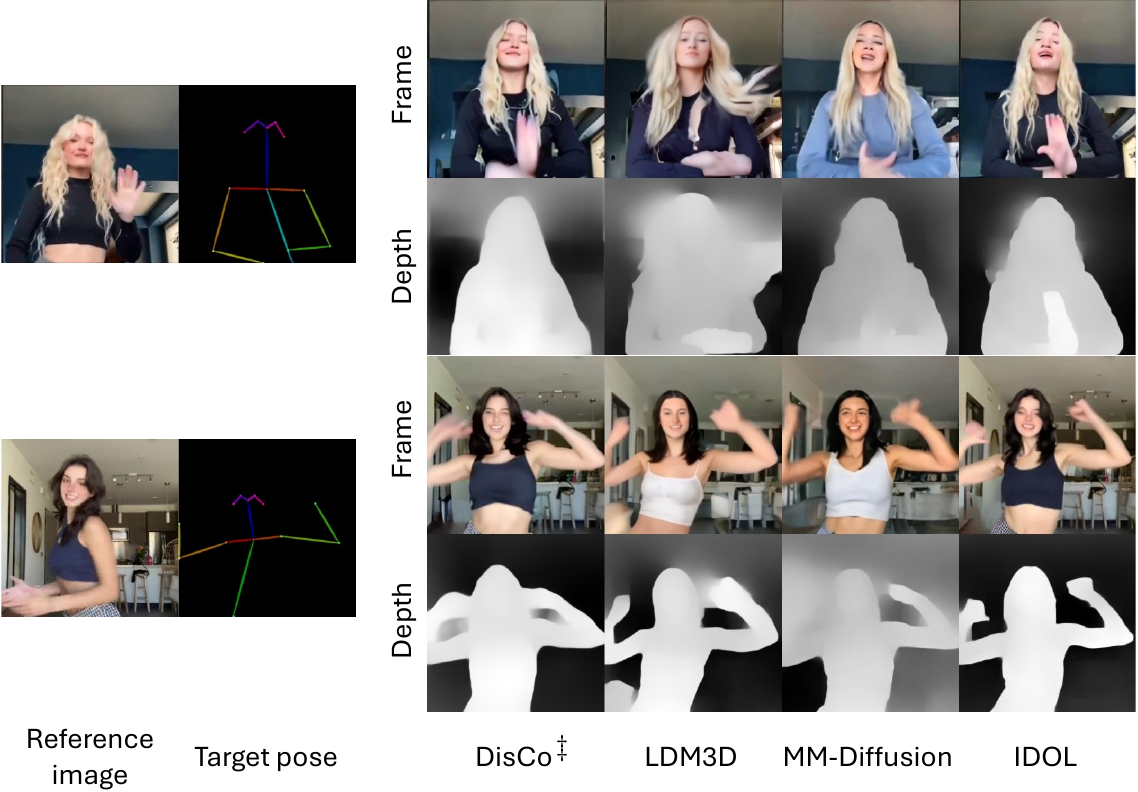}
    \caption{Qualitative results comparison on
    MiDaS~\cite{ranftl2020towards,ranftl2021vision} grayscale whole-frame depth
    map generation.}
    \label{fig:qa-comparison-midas}
\end{minipage}
\end{figure}

%% file: tables/tab-adaptation.tex
\begin{table}[t!]
    \centering
    \resizebox{0.85\linewidth}{!}{%
\begin{tabular}{ccccccc}
\hline 
\multirow{2}{*}{Motion repr. } & \multirow{2}{*}{Temporal modeling} & \multirow{2}{*}{Pose control} & \multicolumn{2}{c}{{Video}} & \multicolumn{1}{c}{{Depth}} & \multicolumn{1}{c}{{Image}}\tabularnewline
 &  &  & {FID-FVD$\downarrow$} & {FVD$\downarrow$} & {L2$\downarrow$} & {FID$\downarrow$}\tabularnewline
\hline 
\rowcolor{mygray} OpenPose~\cite{cao2017realtime} & Handcrafted & ControlNet~\cite{zhang2023adding} & 17.86 & 223.69 & 0.0336 & 36.04\tabularnewline
OpenPose~\cite{cao2017realtime} & AnimateDiff~\cite{guo2023animatediff} & ControlNet~\cite{zhang2023adding} & 19.58 & 201.83 & 0.0350 & 33.14 \tabularnewline
DWPose~\cite{yang2023effective} & AnimateDiff~\cite{guo2023animatediff} & ControlNet~\cite{zhang2023adding} & 16.73 & 179.20 & 0.0245 & 31.06\tabularnewline
DWPose~\cite{yang2023effective} & AnimateDiff~\cite{guo2023animatediff} & T2I-Adapter~\cite{mou2023t2i} & 14.40 & 188.16 & 0.0195 & 30.06\tabularnewline
\hline 
\end{tabular}
    }
    \caption{Quantitative result of our \ourmodel generalize to different motion
    representations, temporal modeling designs, and pose control module on
    TikTok~\cite{jafarian2021learning} with HDNet
    depth~\cite{jafarian2021learning}. The \colorbox{mygray}{gray} row indicates
    our default setting. Please find our supplementary material for additional
    implementation details.}
    \label{tab:adaptation}
\end{table}

%% file: figures/comb-fig-qa-ablation-tab-ablation-on-pre-train.tex
\begin{figure}[t]
    \begin{minipage}[h]{0.49\linewidth}
        \includegraphics[width=\linewidth]{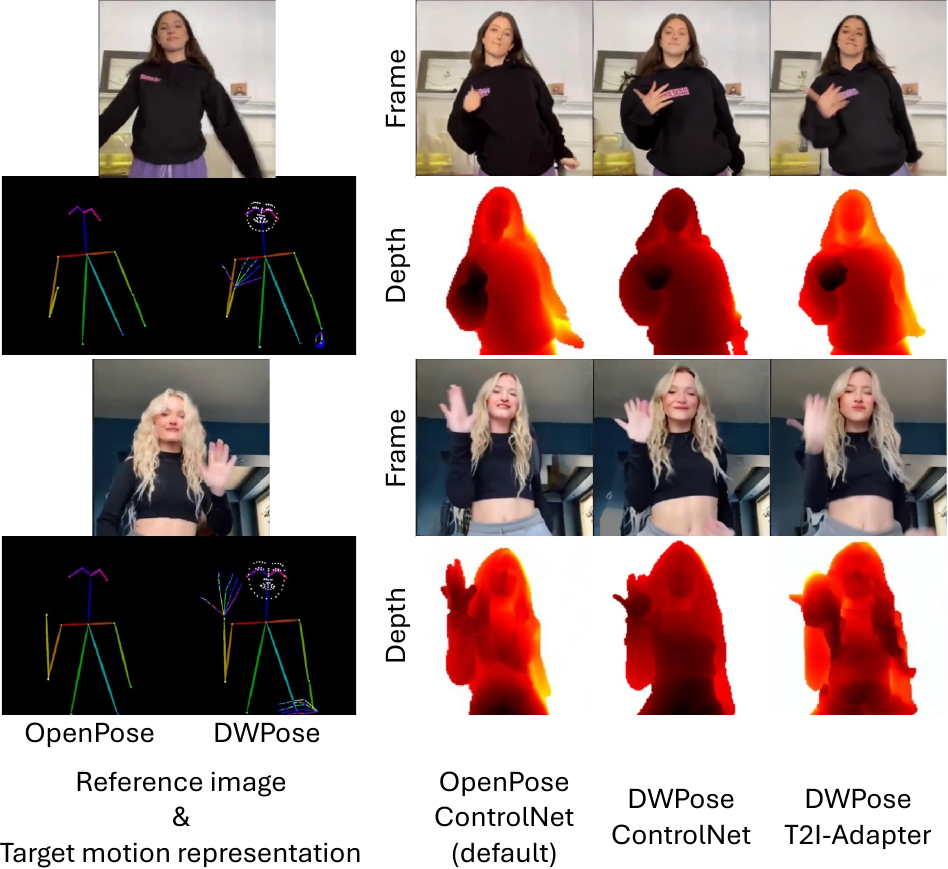}
        \caption{Qualitative results of our \ourmodel generalize to different
        motion representations and pose control modules.}
        \label{fig:adaptation}
    \end{minipage}
    \hfill
    \begin{minipage}[h]{0.49\linewidth}
        \includegraphics[width=\linewidth]{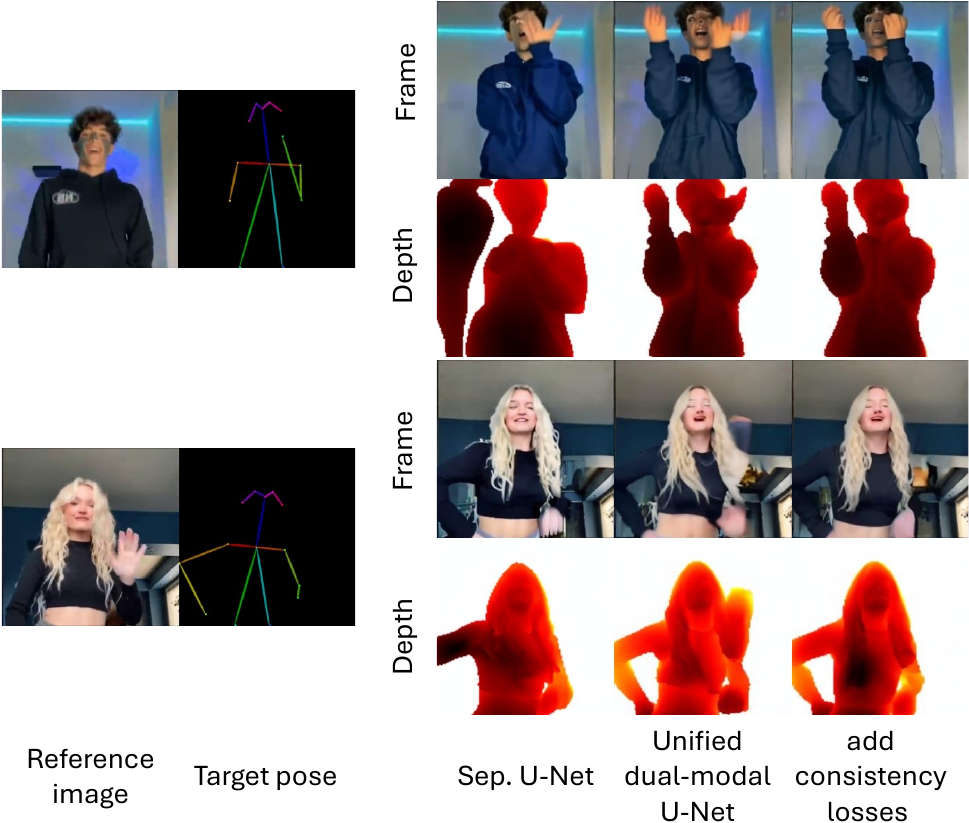}
        \caption{Qualitative comparison between the baseline video LDM, unified dual-modal U-Net w/o and w/ consistency losses.}
        \label{fig:qa-ablation}
    \end{minipage}
\end{figure}

%% file: tables/comb-tab-ablation-on-architecture-ablation-on-loss.tex
\begin{table}[t]
    \begin{minipage}[h]{0.55\linewidth}
        \centering
        \resizebox{\linewidth}{!}{%
        \begin{tabular}{lrcccc}
            \hline 
            \multirow{2}{*}{Settings} & \multirow{2}{*}{\#Param.} & \multicolumn{2}{c}{{Video}} & \multicolumn{1}{c}{{Depth}} & \multicolumn{1}{c}{{Image}}\tabularnewline
             &  & {FID-FVD$\downarrow$} & {FVD$\downarrow$} & {L2$\downarrow$} & {FID$\downarrow$}\tabularnewline
            \hline 
            Sep. U-Net for joint denoise & $2\times1.39$B & 24.28 & 282.50 & 0.0822 & 41.72\tabularnewline
            Share U-Net for joint denoise & $1.39$B & 22.10 & 272.37 & 0.0369 & 39.43\tabularnewline
            + Cross-modal attn. & $1.41$B & \textbf{19.28} & \textbf{260.65} & \textbf{0.0360} & \textbf{39.01}\tabularnewline
            \hline 
        \end{tabular}%
        }
        \caption{Ablation study on the joint video-depth learning and the unified
        dual-modal U-Net design.}
        \label{tab:ablation-on-architecture}
    \end{minipage}
    \hfill
    \begin{minipage}[h]{0.43\linewidth}
        \centering
        \resizebox{0.8\linewidth}{!}{%
        \begin{tabular}{cccccc}
        \hline 
        \multirow{2}{*}{$\mathcal{L}_{\text{xattn}}$} & \multirow{2}{*}{$\mathcal{L}_{\text{mo}}$} & \multicolumn{2}{c}{{Video}} & \multicolumn{1}{c}{{Depth}} & \multicolumn{1}{c}{{Image}}\tabularnewline
         &  & {FID-FVD$\downarrow$} & {FVD$\downarrow$} & {L2$\downarrow$} & {FID$\downarrow$}\tabularnewline
        \hline 
         &  & 19.28 & 260.65 & 0.0360 & 39.01\tabularnewline
        $\checkmark$ &  & 19.99 & 244.58 & 0.0351 & 37.89\tabularnewline
        $\checkmark$ & $\checkmark$ & \textbf{17.86} & \textbf{223.69} & \textbf{0.0336} & \textbf{36.04}\tabularnewline
        \hline 
        \end{tabular}%
        }
        \caption{Ablation study on the video-depth consistency loss functions
        $\mathcal{L}_{\text{mo}}$ and $\mathcal{L}_{\text{xattn}}$.}
        \label{tab:ablation-on-losses}
    \end{minipage}
\end{table}

%% file: tables/comb-tab-ablation-on-pretrain-complexity.tex
\begin{table}[t]
    \begin{minipage}[h]{0.43\linewidth}
        \centering
        \resizebox{\linewidth}{!}{%
        \begin{tabular}{ccccc}
            \hline 
            \multirow{2}{*}{Pre-training strategy} & \multicolumn{2}{c}{{Video}} & \multicolumn{1}{c}{{Depth}} & \multicolumn{1}{c}{{Image}}\tabularnewline
             & {FID-FVD$\downarrow$} & {FVD$\downarrow$} & {L2$\downarrow$} & {FID$\downarrow$}\tabularnewline
            \hline 
            w/o pre-training & 39.24 & 452.81 & 0.0363 & 56.80\tabularnewline
            HAP~\cite{wang2023disco} & 19.83 & 227.86 & \textbf{0.0336} & 38.71\tabularnewline
            HAOP (ours) & \textbf{17.86} & \textbf{223.69} & \textbf{0.0336} & \textbf{36.04}\tabularnewline
            \hline 
        \end{tabular}%
        }
        \caption{Ablation study on the pre-training.}
        \label{tab:comp-on-pre-train}
    \end{minipage}
    \hfill
    \begin{minipage}[h]{0.55\linewidth}
        \centering
        \resizebox{\linewidth}{!}{%
        \begin{tabular}{ccccc}
        \toprule
        Method & FLOPs (T) & \#Param. (B) & GPU mem. (MB) & Infer. time (s) \\
        \midrule
        LDM3D~\cite{stan2023ldm3d} & 48.80 & 1.61 & \textbf{7661} & 42.02 \\
        MM-Diffusion~\cite{ruan2023mm} & 41.02 & 2.78 & 14651 & 13.49 \\
        \rowcolor{mygray} \ourmodel (ours) & \textbf{39.35} & \textbf{1.41} & 10251 & \textbf{12.23} \\
        \bottomrule
        \end{tabular}%
        }
        \caption{Computational complexity comparison.}
        \label{tab:complexity}
    \end{minipage}
\end{table}

%% file: sec/5_conclusion.tex
\section{Conclusion}
\label{sec:conclusion}

In this paper, we propose \ourmodel, a framework tailed for human-centric joint 
video-depth generation.
Our proposed unified dual-modal U-Net improves the video and depth synthesis by
implicit video structure learning, with cross-modal attention explicitly bridging
the joint video-depth denoising process.
Our motion consistency loss and cross-attention map consistency loss promote
spatial alignment between the generated video and depth.
Extensive experiments on the TikTok and NTU120 datasets show our superior
performance compared with existing methods, and the adaption ability to
different kinds of depth maps.
Our \ourmodel is also able to generalize to different motion representations and
pose control modules.

\noindent\textbf{Limitations.}
Despite the performance advantage of our \ourmodel, it faces several major
limitations.
First, the computational demands of processing dual-modal data, particularly at
high resolutions, hinder its suitability for real-time applications,
highlighting a need for further optimization.
Additionally, the reliance on high-quality depth maps for training limits its
applicability in scenarios where such data is limited or of low quality.
Addressing this, future work may explore unsupervised methods or data
augmentation strategies to mitigate the data quality constraint.

\noindent\textbf{Negative societal impact.}
Our model raises ethical concerns, including the potential for creating deepfake
videos, producing biased outputs, and threatening intellectual property rights.
To mitigate these risks, we can incorporate invisible watermarks to ensure
content authenticity.

%% file: sec/6_acknowledgement.tex
\section*{Acknowledgements}

This work is supported in part by the Defense Advanced Research Projects Agency (DARPA) under Contract No.~HR001120C0124. Any opinions, findings and conclusions or recommendations expressed in this material are those of the author(s) and do not necessarily reflect the views of the Defense Advanced Research Projects Agency (DARPA).

%% file: sec/1_exp.tex
\section{Additional experiments}

\noindent

\noindent\textbf{Qualitative results.}
Please refer to the accompanying video for the qualitative comparison.
In the video demonstration, we stitch each generated video snippets together to
form a long video, which may lead to unnatural transitions between each snippet.
We make the following observations.
(1)~Compared with DisCo~\cite{wang2023disco}, LDM3D~\cite{stan2023ldm3d}, and
MM-Diffusion~\cite{ruan2023mm}, our generated videos and depth sequences exhibit
more natural and smoother transitions, demonstrating the effectiveness of our
method.
A notable observation on the NTU120 dataset~\cite{shahroudy2016ntu,liu2019ntu}
is LDM3D's limitation in generating diverse frames after fine-tuning the
autoencoder, often resulting in repetitive ``stuck'' video sequences. 
This issue may stem from the subtle motions and predominant static content in
NTU120 videos, suggesting that fine-tuning the autoencoder might necessitate a
larger and more varied training dataset.
(2)~Our \ourmodel is able to composite different foreground and background,
while simultaneously generating video and depth.
This feature distinguishes it from concurrent
methods~\cite{xu2023magicanimate,chang2023magicdance,hu2023animate}, offering a
unique capability in the area of video-depth synthesis.
(3)~Our method is able to generalize to different pose conditions, such as
OpenPose~\cite{cao2017realtime} and DWPose~\cite{yang2023effective}.

\noindent\textbf{Cross-attention map consistency.}
To enhance video-depth alignment in \ourmodel, we propose a cross-attention map
consistency loss $\mathcal{L}_{\text{xattn}}$.
This loss function encourages alignment of the video and depth
cross-attention maps.
We also explore alternate ways to align the cross-attention maps.
One straightforward approach is to use a shared cross-attention map for both
branches.
We test two variations: replacing individual cross-attention maps with their
average, and using the video stream's cross-attention map as a substitute for
both.
Our results in~\cref{tab:ablation-on-xattn} reveal that sharing a
cross-attention map significantly reduces performance, particularly impacting
depth L2 accuracy (as seen in rows 2 and 3).
These findings highlight the need for each stream to maintain diverse
cross-attention maps to produce high-quality outputs.
Our implementation of $\mathcal{L}_{\text{xattn}}$ effectively balances the need
for consistency with the preservation of each map's unique characteristics,
ultimately contributing to superior overall performance.

\input{tables/tab-ablation-on-xattn.tex}

%% file: tables/tab-ablation-on-xattn.tex
\begin{table}[h!]
\centering
\resizebox{0.7\linewidth}{!}{%
\begin{tabular}{lcccc}
    \hline 
    \multirow{2}{*}{Setting} & \multicolumn{2}{c}{{Video}} & \multicolumn{1}{c}{{Depth}} & \multicolumn{1}{c}{{Image}}\tabularnewline
     & {FID-FVD$\downarrow$} & {FVD$\downarrow$} & {L2$\downarrow$} & {FID$\downarrow$}\tabularnewline
    \hline 
    - & 19.28 & 260.65 & 0.0360 & 39.01\tabularnewline
    Share cross-attention map (avg) & 20.82 & 297.76 & 0.0706 & 49.66\tabularnewline
    Share cross-attention map (video) & 20.00 & 253.73 & 0.0718 & 44.58\tabularnewline
    Apply $\mathcal{L}_{\text{xattn}}$ & \textbf{19.99} & \textbf{244.58} & \textbf{0.0351} & \textbf{37.89}\tabularnewline
    \hline 
\end{tabular}%
}
\caption{Ablation study on the cross-attention map operations on the TikTok
dataset~\cite{jafarian2021learning} with HDNet
depth~\cite{jafarian2021learning}.}
\label{tab:ablation-on-xattn}
\end{table}

%% file: sec/2_details.tex
\section{Implementation details}

Our code is developed based on diffusers~\cite{von-platen-etal-2022-diffusers}.
We follow DisCo~\cite{wang2023disco} to use Stable Diffusion
v1.4~\cite{rombach2022high} as the backbone.
For HAOP pre-training, we follow DisCo~\cite{wang2023disco} to freeze the
ResBlocks and train the model for 25k steps, with input image size
$256\times 256$ and learning rate $1e^{-3}$.
For fine-tuning, we adopt a two-stage approach.
In the first stage, the temporal layers are removed,
and the framework is trained on joint image-depth denoising.
In the second stage, the whole framework with temporal modules is trained for
the joint video-depth denoising.
Both the first and the second stages are trained for 15k steps with a learning
rate of $1e^{-4}$.
For the second stage, the model is trained on 8-frame sequences.
Both the pre-training and fine-tuning are conducted on 32 V100 GPUs.
The weight hyper-parameters are set via a grid search:
$w_\text{mo}=w_\text{xattn}=0.01$.
We set the temperature term $\tau$ in the motion field computation to $1 / \sqrt{D_n}$, where $D_n$ is the number of channels in the $n$-th layer.

\noindent\textbf{Comparison methods.}
We use DisCo~\cite{wang2023disco}, a recent diffusion-based human dance video
generation method, as a strong human-centric video generation
baseline.
As a pioneering method directly tailed for human-centric joint video-depth
generation, we compare our \ourmodel with the closest multi-modal generation
counterparts.
We choose MM-Diffusion~\cite{ruan2023mm}, initially designed for
text-to-video-audio synthesis, and LDM3D~\cite{stan2023ldm3d}, aimed at
text-to-image-depth generation.
To facilitate a fair comparison, we align their backbones to the same video LDM
baseline, adapting them for the human-centric video-depth task.
For MM-Diffusion~\cite{ruan2023mm}, we replace the audio U-Net with a duplicate
of the video U-Net (without sharing parameters, unlike in our \ourmodel) and
retain the rest of the structure unchanged.
In the case of LDM3D~\cite{stan2023ldm3d}, we inflate the 2D U-Net to a 3D U-Net
to accommodate video generation.
Both adapted methods employ ControlNet~\cite{zhang2023adding} for human pose
control and process background and foreground inputs similarly to \ourmodel,
ensuring consistency in our comparative evaluation.

\noindent\textbf{Generalization to different designs.}
In our main manuscript, we evaluated the generalization ability of our \ourmodel to
different designs, including DWPose~\cite{yang2023effective},
AnimateDiff~\cite{guo2023animatediff}, and T2I-Adapter~\cite{mou2023t2i}.
For the adaptation of AnimateDiff~\cite{guo2023animatediff}, we remove the
original temporal convolutional and attention layers in the 3D U-Net, and insert
the AnimateDiff~\cite{guo2023animatediff} pre-trained motion modules.
For the T2I-Adapter~\cite{mou2023t2i}, we replace the original pose ControlNet
with a pre-trained OpenPose T2I-Adapter.
Note that the video and depth streams share the same pose T2I-Adapter, similar
to the pose ControlNet.